\documentclass[conference]{IEEEtran}
\usepackage{cite}
\usepackage{amsmath,amssymb}
\usepackage{algorithmic}
\usepackage{graphicx}
\usepackage{textcomp}
\usepackage{multirow}
\usepackage{hyperref}
\usepackage{booktabs}

\usepackage{xcolor}

\DeclareRobustCommand*{\IEEEauthorrefmark}[1]{%
  \raisebox{0pt}[0pt][0pt]{\textsuperscript{\footnotesize #1}}%
}

\begin{document}

\title{Neural Operator Learning for Long-Time Integration in Dynamical Systems with Recurrent Neural Networks\\
}

\author{\IEEEauthorblockN{Katarzyna Michałowska\IEEEauthorrefmark{1}\IEEEauthorrefmark{2}\IEEEauthorrefmark{3} \hspace{9pt} Somdatta Goswami\IEEEauthorrefmark{3} \hspace{9pt} George Em Karniadakis\IEEEauthorrefmark{3}\IEEEauthorrefmark{4} \hspace{9pt} Signe Riemer-Sørensen\IEEEauthorrefmark{1}}
\vspace{10pt}
\IEEEauthorblockA{
\IEEEauthorrefmark{1}Department of Mathematics and Cybernetics, SINTEF Digital, Oslo, Norway}
\IEEEauthorblockA{
\IEEEauthorrefmark{2}Department of Physics, University of Oslo, Oslo, Norway}
\IEEEauthorblockA{
\IEEEauthorrefmark{3}Department of Applied Mathematics, Brown University, Providence, Rhode Island, USA}
\IEEEauthorblockA{
\IEEEauthorrefmark{4}School of Engineering, Brown University, Providence, Rhode Island, USA}

}

\maketitle

\begin{abstract}
Deep neural networks are an attractive alternative for simulating complex dynamical systems, as in comparison to traditional scientific computing methods, they offer reduced computational costs during inference and can be trained directly from observational data. 
Existing methods, however, cannot extrapolate accurately and are prone to error accumulation in long-time integration.  
Herein, we address this issue by combining neural operators with recurrent neural networks, learning the operator mapping, while offering a recurrent structure to capture temporal dependencies.
The integrated framework is shown to stabilize the solution and reduce error accumulation for both interpolation and extrapolation of the Korteweg-de Vries equation. 

\end{abstract}


\section{Introduction}
Dynamical systems modeling is formally concerned with the analysis, forecasting, and understanding of the behavior of ordinary or partial differential equation (ODE/PDE) systems or similar iterative mappings that represent the evolution of a system's state. Modern machine learning methods have opened up a new area for building fast emulators for solving parametric ODEs and PDEs. 
One class of such frameworks are neural operators, which have been gaining popularity as surrogate models for dynamical systems in recent years \cite{Lu_2021, goswami2022neural, goswami2022deep}. While previous research on neural architectures has primarily focused on learning mappings between finite-dimensional Euclidean spaces \cite{raissi2019physics,samaniego2020energy}, neural operators learn mappings between infinite-dimensional Banach spaces \cite{goswami2022physics}. 

The two popular neural operators which have shown promising results so far are the deep operator network (DeepONet) \cite{Lu_2021} introduced in 2019 and the Fourier Neural operator (FNO) \cite{fourier_2020} introduced in 2020. Both operator networks have been applied to solving complex problems for real-world applications in diverse scientific domains, including medicine \cite{goswami2022neural}, physics \cite{lin2021operator, wen2022u}, climate \cite{kissas2022learning, bora2023learning, pathak2022fourcastnet} and materials science \cite{goswami2022114587, you2022learning, rashid2022learning}. Regardless of the impressive results presented in these works, the problem of approximating system's behavior over a long-time horizon employing neural operators remains underexplored \cite{goswami2023learning, wang2023long}.

To address the issues of operator learning over a long-time horizon, some recent works suggest employing physics-informed DeepONets \cite{wang2023long}, using transfer learning and later fine-tuning the pre-trained DeepONet with sparse measurements in the extrapolated zone or with a PDE loss \cite{zhu2022reliable}, and a hybrid inference approach, integrating neural operators with high-fidelity solvers \cite{oommen2022learning}. These approaches, however, require either the knowledge of the underlying PDE or sparse labels in extrapolation. 

In this work, we build upon \cite{michalowska2023don}, where we introduce a multi-resolution approach combining DeepONets with long short-term memory networks. The aforementioned method leverages the discretization-invariance property of DeepONets to extend the training sample by low-resolution data, effectively allowing the model to learn from a larger and more representative training sample. In this study, on the other hand, we examine the following: a) We explore the combinations of the DeepONets and FNOs with alternative recurrent architectures (basic recurrent neural network (RNN), gated recurrent unit (GRU), and long short-term memory network (LSTM)) on single-resolution data, and b) we examine the performance of the proposed combinations in extrapolation. Our tests show that these neural operator extensions can significantly improve the prediction for long-time horizons in both interpolation and extrapolation tasks.


\section{Recurrent networks with neural operators}

\begin{figure*}[ht]
\begin{center}
\includegraphics[width=0.95\textwidth]{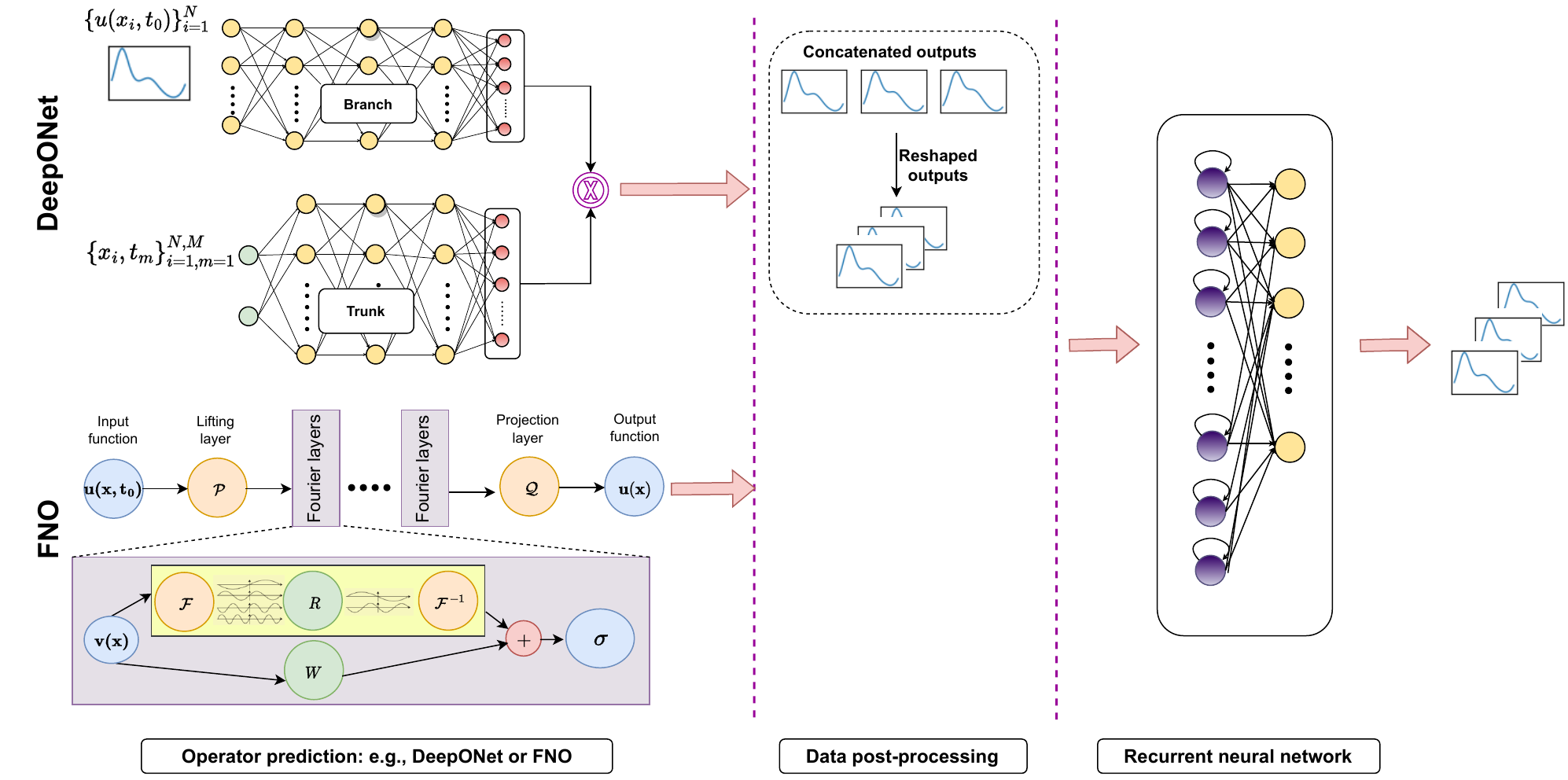}
\caption{The architecture combining a deep neural operator (e.g., DeepONet or FNO) and a recurrent neural network architecture (Figure adapted from \cite{michalowska2023don}). The neural operator learns the mapping from the initial condition $u_{t=0}$ to solutions at later timesteps. These solutions are then represented as a sequence and fed into an RNN, which outputs the solution.}
\label{fig:proposed_architecture}
\end{center}
\end{figure*}

We have considered the two most studied neural operators: the deep operator network (DeepONet) and the Fourier neural operator (FNO), combined with one of the three most successful recurrent neural network: the traditional RNN (non-gated), LSTM, and GRU. The full architecture is set up such that the outputs of the neural operator are fed as inputs to the RNN. The neural operator learns the solution operator, while the RNN processes the outputs of the neural operator as temporal sequences. A schematic representation of the approach is shown in \autoref{fig:proposed_architecture}. 

The aim of the neural operator is to learn the mapping between two infinite-dimensional functional spaces, where we learn the evolution of a dynamical system from a given initial condition. Such operator mapping can be represented as:
$$\mathcal{N}: u(x,t=0) \rightarrow [u(x, t_1), ..., u(x, t_n)],$$
where $n$ is the total number of temporal discretization points that defines the full trajectory and $\mathcal N$ is the nonlinear operator that defines the PDE. The RNN is employed to identify the outputs of the operator as a sequence. The work is motivated by the fact that in feed-forward neural networks (including neural operators) all the outputs are independent, which is not true for sequence learning. Recurrent networks address this issue through feedback loops and hidden states which allow for the information to be passed forward and capture the temporal dependencies between the outputs. 


\subsection{Neural operator learning}
\label{deep_operator_explanations}

Neural operators learn nonlinear mappings between infinite-dimensional functional spaces on bounded domains, providing a unique simulation framework for the real-time prediction of multi-dimensional complex dynamics. Once trained, such models are discretization invariant, which means they share the same network parameters across different parameterizations of the underlying functional data. In this study, we consider the performance of two neural operators, DeepONet proposed in \cite{Lu_2021} and FNO proposed in \cite{fourier_2020} with our novel recurrent-networks-integrated neural operator architecture (\autoref{fig:proposed_architecture}).

\noindent\textbf{FNO}: The Fourier Neural Operator is based on Green's theorem, which involves parameterizing the integral kernel in the Fourier space. The input to the network is elevated to a higher dimension, then passed through several Fourier layers before being projected back to the original dimension with a feed-forward layer. Each Fourier layer involves a forward fast Fourier transform (FFT), followed by a linear transformation of the low-Fourier modes and then an inverse FFT. Finally, the output is added to a weight matrix, and the sum is passed through an activation function to introduce nonlinearity. Different variants of FNO have been proposed based on the pre-decided dimensions to replace the integral kernel with the convolution operator defined in Fourier space. A numerical problem that is $1$D in space and $1$D in time could be handled using either FNO-$2$D, which employs Fourier convolutions through space and time to learn the dynamics directly over multiple timesteps, given an initial condition, or using FNO-$1$D, which performs Fourier convolution in space and uses a recurrent time-marching approach to propagate the solution in time. In this work, we employ FNO-$2$D, since time-marching approaches are known to be prone to error accumulation in long time series prediction.


\noindent\textbf{DeepONet}: The deep operator network is based on the universal approximation theorem for operators \cite{chen1995universal} and employs two deep neural networks (branch and trunk network) to learn a family of PDEs and provide a discretization-invariant emulator, which allows for fast inference and low generalization error \cite{lanthaler2022error}. The branch network encodes the input function (the initial or boundary conditions, constant or variable coefficients, source terms) at fixed sensor points while the trunk network encodes the information related to the spatio-temporal coordinates of the output function. The output embeddings of the branch and the trunk networks are multiplied element-wise and summed over the neurons in the last layer of the networks. Once the network is trained, it can be employed to interpolate the solution at any spatial and temporal location within the domain.


\subsection{The choice of the recurrent neural network}
\label{RNN_explanations}

Feed-forward neural networks lack explicit mechanisms to learn dependencies between outputs, which is a fundamental problem when learning temporal sequences. Recurrent neural networks are an extension over traditional neural networks designed to improve sequence learning by preserving a hidden state which carries information from the previous time steps. During training, this hidden state is updated with the current input and the previous hidden state. Another way of learning temporal sequences is through autoregressive methods, which recursively update the input function using the outputs obtained at the previous time steps. This approach, however, is known to result in high error accumulation when compared to direct multi-step prediction methods, such as recurrent neural networks.


Since traditional RNNs are prone to vanishing gradients in long sequences, \textit{i.e.}, the gradients used in training approach zero as they are multiplied with each time step to adjust the weights. Therefore, we also investigate the proposed architecture when combined with GRU and LSTM. GRUs and LSTMs address the problem of vanishing gradients by introducing a set of gates, implemented as Sigmoid functions, that allow or block the flow of long-term information through the network. Since the name RNN can refer to both the whole class of recurrent neural networks, as well as the specific architecture, we further use the term \textit{simple RNN} when referring to non-gated recurrent architecture and \textit{RNN} for all types of recurrent neural networks, including simple RNN, GRU, and LSTM. 


\subsection{Simultaneous vs. two-step training}
\label{subsec:training_process}
We propose two modes of training the networks in the combined architectures: simultaneous training, in which at each training step the weights of both networks are updated in a single backward pass, and two-step training, where we first train the neural operator network and then the RNN. 

The two-step approach allows to take advantage of the resolution-invariance property of neural operators in the training.
While the RNNs require evenly spaced data, the operator alone can be trained in a discretization-invariant manner, \textit{e.g.}, to increase the number of training samples when multi-resolution data is available.
The neural operator is trained until its validation error does not change substantially in consecutive epochs and is then used in inference to produce the training data for the RNN. The RNN learns a mapping between the outputs of the neural operator and the ground truth, in a manner equivalent to freezing the weights of the neural operator network and continuing the training. However, this approach can also cause additional error propagation since the approximation error in the neural operator layers cannot be reduced at the point of training the RNN, which is not the case in the simultaneous training mode.

In the remainder of the paper, we refer to the architectures trained in the simultaneous mode as DON-RNN and FNO-RNN, and as DON+RNN, FNO+RNN, and equivalent to the architectures trained in the two-step mode.

\section{Training and testing data}









\subsection{Korteweg de Vries equation}

To evaluate the performance of the proposed variants, we consider the Korteweg-de Vries (KdV) equation, defined as:
\begin{equation}
\label{eq:kdv}
    u_t  - \eta uu_x + \gamma u_{xxx} = 0,
\end{equation}
where $u$ is the amplitude of the wave, $\eta$ and $\gamma$ are chosen real-valued scalar parameters, $x$ is a spatial and $t$ is the time dimension. 

The chosen initial condition, $u(x,0)$ is a sum of two solitons, \textit{i.e.}, $u=u_1+u_2$. A single soliton is a type of solitary wave and is expressed as:
\begin{equation}
{u_i(x,0) =  2 k_i^2  \text{sech}^2 \left( k_i ( (x+\frac{P}{2}-Pd_i) \% P - \frac{P}{2} )\right)^2,}
\end{equation}
where $\mathrm{sech}$ stands for hyperbolic secant ($\frac{1}{\mathrm{cosh}}$), $P$ is the period in space, $\%$ is a modulo operator, $i=\{1,2\}$, $k\in [0.5, 1.0]$ and $d\in [0, 1]$ are coefficients that determine the height and location of the peak of a soliton, respectively. 

To train the recurrent-networks-integrated neural operator, we generated $N = 5{,}000$ initial conditions, $u(x,t=0)$, and the evolution dynamics was modeled using the midpoint method. The simulation takes place on a $1$D domain $\Omega= [0,10]$  discretized with $50$ uniformly spaced grid points ($\Delta x=0.2$) in the time interval $t=[0,5]$ for $\Delta t=0.025$, thereby discretizing the temporal domain into $201$ points. To form the training data set, we collect solution snapshots $u(x,t)$ at $n_t=200$. 

The generated labeled dataset with $N$ unique realizations is split into training and testing sets such that the number of training samples, $N_{train} = 0.9\times N$ and the number of testing samples, $N_{test} = 0.1\times N$. Furthermore, $10$\% of the training set is used explicitly for validation during the training process. 

\begin{figure*}[t]
    \centering
    \begin{minipage}{0.95\columnwidth}
        \centering
        \includegraphics[width=\linewidth]{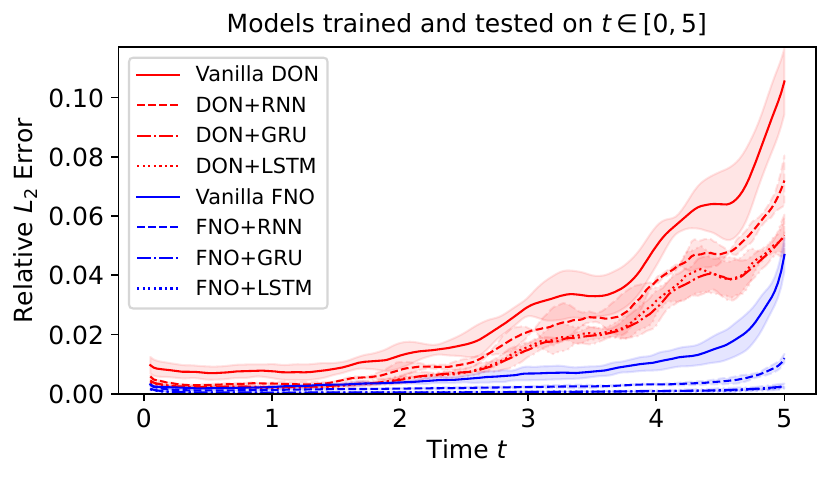} 
        \small{(a) Interpolation, two-step training.}
    \end{minipage}\hspace{1cm} 
    \begin{minipage}{0.95\columnwidth}
        \centering
        \includegraphics[width=\linewidth]{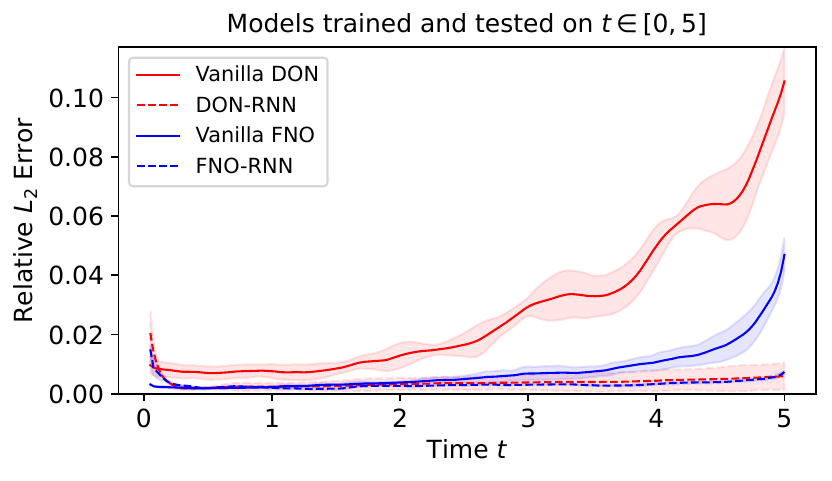} 
        \small{(b) Interpolation, simultaneous training.}
    \end{minipage}
    \caption{E$1$: Interpolation performance. The models are both trained and tested on full trajectories ($200$ steps in $t\in[0,5]$). The error is given as the relative squared error over all spatial points $x$ for each time step $t$ on the test data. The enhanced models are compared to vanilla neural operators. The shaded area indicates the 95\% confidence interval calculated over the three trained models (two standard deviations of the error).}
\label{fig:interpolation_performance}
\end{figure*}


\begin{figure*}[t]
    \centering
    \begin{minipage}{0.95\columnwidth}
        \centering
        \includegraphics[width=\linewidth]{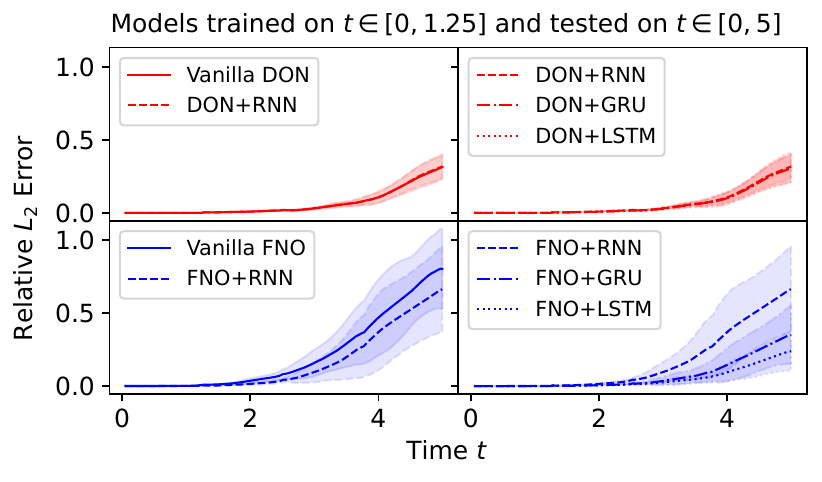} 
        \small{(a) Extrapolation, two-step training.}
    \end{minipage}\hspace{1cm} 
    \begin{minipage}{0.95\columnwidth}
        \centering
        \includegraphics[width=\linewidth]{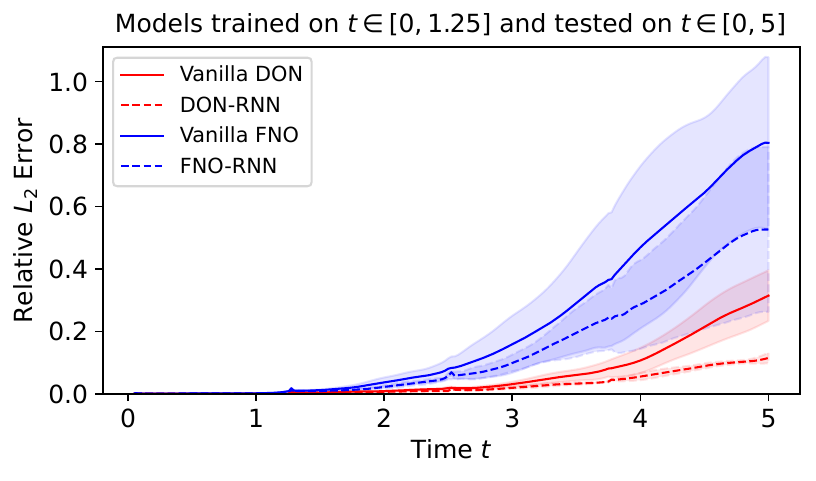} 
        \small{(b) Extrapolation, simultaneous training.}
    \end{minipage}
    \caption{E$2$: Extrapolation performance. The models are trained on partial trajectories ($50$ steps in $t\in[0,1.25]$) and tested on full trajectories ($200$ steps in $t\in [0,5]$). The error is given as the relative squared error over all spatial points $x$ for each time step $t$ on the test data. The panels show different combinations of vanilla operators and enhanced models.  The shaded area indicates the 95\% confidence interval calculated over the three trained models (two standard deviations of the error).}
\label{fig:extrapolation_performance}
\end{figure*}

\section{Problem statement}

To test the proposed variants of the neural operator and recurrent neural network architectures, we carried out the following two experiments:
\begin{itemize}
 \item \textbf{E1: }Training and testing on the mapping $\mathcal N: u(x, t=0) \rightarrow u(x,t)$, where $t\in [0.025, 5]$, where $\mathcal N$ denotes the integrated model.
 \item \textbf{E2: }Training to learn the mapping $\mathcal N: u(x, t=0) \rightarrow u(x,t)$, where $t\in [0, 1.25]$, and testing on $t\in [1.25, 5]$ by recursively updating the the initial condition.
\end{itemize}

For each of the experiments, we have considered the two training processes discussed in \autoref{subsec:training_process}, and have carried out an analysis for the combination of two neural operators (DeepONet and FNO) and recurrent networks (simple RNN, LSTM, and GRU). Furthermore, we have investigated the improvement in the accuracy of the framework against the accuracy of vanilla neural operators.

\section{Results and discussion}
\label{results}

\subsection{E1: Interpolation performance}

The aim of this experiment is to learn the dynamics of the system defined over a time $t \in [0.025, 5]$, which is discretized with $200$ temporal points from the given initial condition. The models are trained and tested on complete trajectories ($200$ steps in time $t\in [0.025, 5]$). In \autoref{metrics_200ts}, we provide the relative $L_1$ error (relative absolute error) and the relative $L_2$ error (relative squared error) on the $N_{test}$ test cases for a total of $10$ different architectures. Overall, the integration of recurrent networks with neural operators improves the accuracy and stability of prediction for all architectures. The predictive error over time for all the architectures is presented in \autoref{fig:interpolation_performance}.

In particular, for both neural operators, the lowest errors in the two-step training are obtained by the LSTM and GRU extensions (see \autoref{metrics_200ts}). While the prediction over all timesteps is obtained in a single forward pass, we still observe an increase in error over time. This can be explained by the solution at later timesteps gradually diverging from the initial condition, provided as input. Notably, the growth rate of the error is significantly reduced after integrating the recurrent networks with neural operators (see \autoref{fig:interpolation_performance}a). The simultaneous training mode shows an advantage over the two-step training for both operators (see \autoref{fig:interpolation_performance}b) as the accumulated error is drastically reduced, which eventually reduces the slope of the error growth. We report a slight increase in the overall error of the simultaneous training of the RNN extension of the FNO architecture (FNO-RNN) compared to the two-step training (FNO+RNN) as seen in \autoref{metrics_200ts}. However, this increase is due to the high error in the first few time steps (\autoref{fig:interpolation_performance}b). 

\begin{figure*}[t]
\vskip 0.2in
\begin{center}
\includegraphics[width=0.9\textwidth]{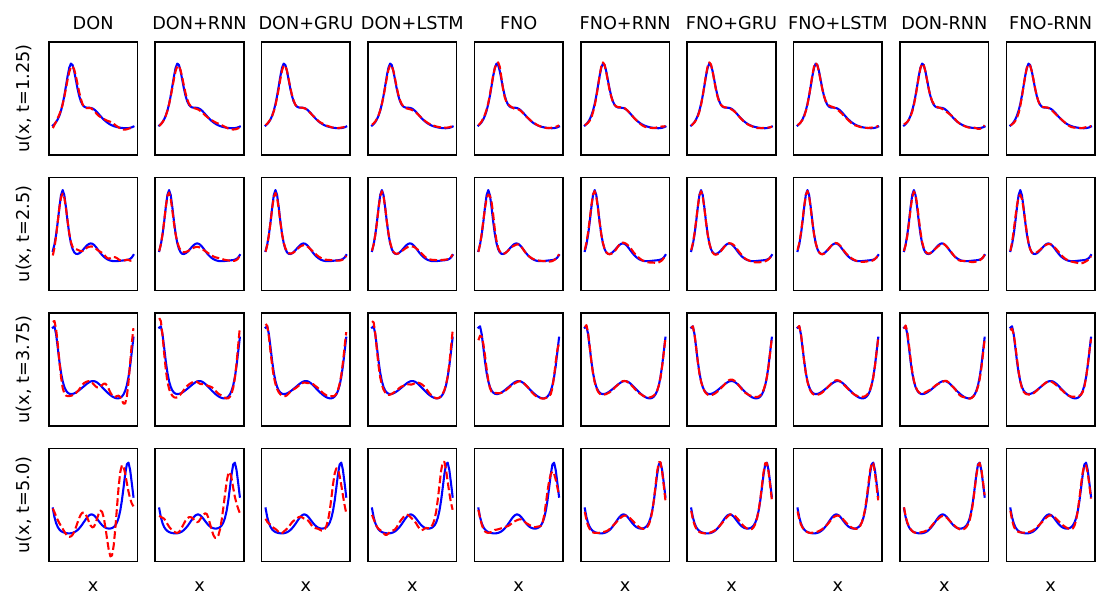}
\caption{E$1$: Interpolation performance on one representative sample for all models trained on $t\in[0,5]$. The full trajectory $t\in[0.025,5]$ is predicted in one shot. Each subplot shows predictions (red dashed line) and the ground truth (blue solid line) row-wise at $t=\{1.25, 2.5, 3.75, 5\}$ and column-wise for all models: vanilla DeepONet, DeepONet+RNN, GRU, LSTM trained in two steps, vanilla FNO, FNO+RNN, GRU, LSTM trained in two steps, and simultaneously trained DON-RNN and FNO-RNN.}
\label{fig:200ts}
\end{center}
\vskip -0.2in
\end{figure*}

For illustration purposes, the prediction plots for all the architectures are shown for four snapshots at $t=\{1.25,2.5,3.75,5\}$ from a single initial condition in \autoref{fig:200ts}. Vanilla DeepONet introduces wiggles to the solution already at time $t=2.5$, and it can be seen that the RNN extensions reduce these artifacts to the point where they disappear completely when the DeepONet is trained simultaneously with a simple RNN. FNO does not seem to suffer from similar instabilities, although it is noticeable that the model does not fit well at the wave peaks in later timesteps. This problem is not visible in the predictions by the extended FNOs.

\subsection{E2: Extrapolation performance}

This experiment aims to test the performance of the integrated framework for extrapolation. To that end, the models are trained to lean the mapping $\mathcal N: u(x, t=0) \rightarrow u(x,t)$, where $t\in [0.025, 1.25]$ is discretized to $50$ temporal points. The model is tested to predict the dynamics over $t\in [0, 5]$ for a given initial condition. During testing, the prediction is performed in a recursive manner, where $50$ time steps are predicted in one shot, and in each iteration, the last predicted observation becomes an input (initial condition) for the next $50$ time steps.  

The error metrics for all the architectures are presented in \autoref{metrics_50ts}. Similarly to the results obtained in the experiment E$1$, we observe lower error for the RNN extension of the neural operator in all cases. 
Furthermore, gated recurrent networks like GRU and LSTM show advantages over simple RNN. \autoref{fig:extrapolation_performance} shows that all models suffer from significant error increase over time. 

In all cases, the neural operators integrated with recurrent networks trained in a two-step procedure show lower error over time, especially for GRU and LSTM (\autoref{fig:extrapolation_performance}a). Significant improvement is observed in the accuracy of FNO integrated with the recurrent networks, where the slope of the error growth is also reduced. However, we observe that the Deeponet+RNN extension performed in the two-step training has limited benefit in extrapolation. For simultaneous training (\autoref{fig:extrapolation_performance}b), the DON-RNN architecture has significantly lower error and slower growth than the vanilla DeepOnet in the last iteration, regardless of a pronounced error increase at the beginning of each iteration (one iteration is a 50-steps-ahead prediction). Furthermore, the FNO-RNN architecture shows similar performance to the vanilla FNO. 

\begin{figure*}[ht]
\vskip 0.2in
\begin{center}
\includegraphics[width=0.9\textwidth]{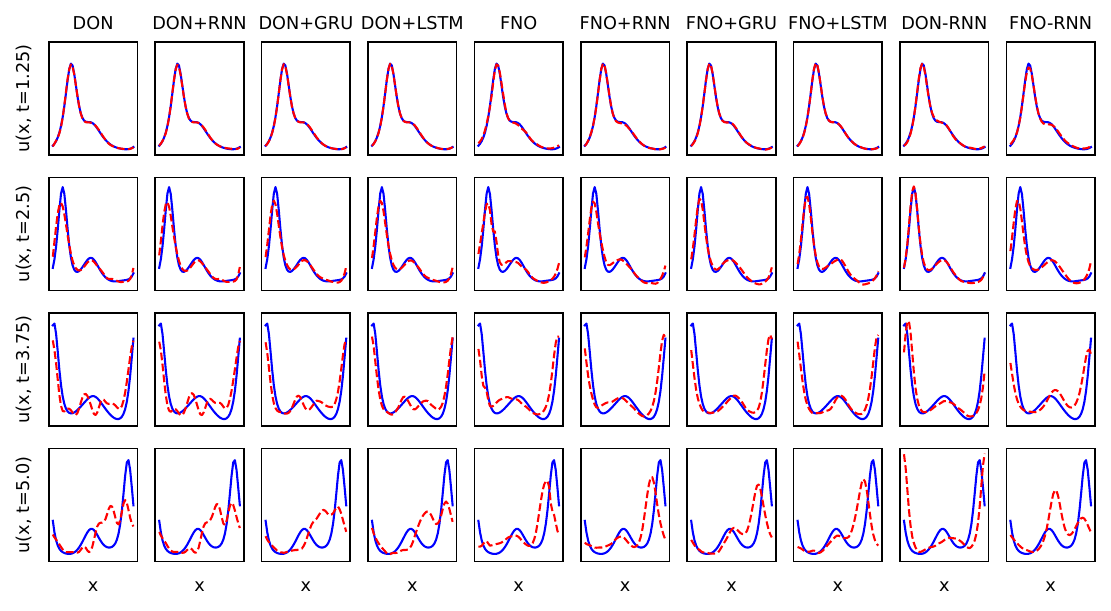}
\caption{E$2$: Extrapolation performance on one representative sample for all models trained on $t\in[0,1.25]$. The full trajectory $t\in[0.025,5]$ is covered recursively as short-interval one-shot predictions, \textit{i.e.}, as $t=0 \rightarrow t\in[0.025,1.25]$, $t=1.25 \rightarrow t\in[1.275,2.5]$, up to $t=5$, where the last predicted solution is used as the initial condition for the next one-shot prediction over the interval $\Delta t=1.25$. Each subplot shows predictions (red dashed line) and the ground truth (blue solid line) for the last solution (the next initial condition) row-wise at $t=\{1.25, 2.5, 3.75, 5\}$ and column-wise for all models: vanilla DeepONet, DeepONet+RNN, GRU, LSTM trained in two steps, vanilla FNO, FNO+RNN, GRU, LSTM trained in two steps, and simultaneously trained DON-RNN and FNO-RNN.}
\label{fig:50ts}
\end{center}
\vskip -0.2in
\end{figure*}

To investigate further, we plot the temporal evolution of a representative sample for four snapshots $t=\{1.25,2.5,3.75,5\}$ in \autoref{fig:50ts}. Overall, it is observed that the models struggle to maintain the wave shape in extrapolation. Each row shows the last time step in one iteration of the recursive prediction. While at $t=1.25$ all models fit the data, at the last time step of the second iteration ($t=2.5$) the peaks of the waves are underestimated, with an exception of simultaneous training of the DON-RNN. At $t=3.75$ the shape of the wave is significantly perturbed for all models but for the two-step training of FNO with GRU and LSTM and the simultaneous training of DON-RNN. Finally, at $t=5.0$ the shape remains partially captured for these models, but with an offset in the spatial dimension, while the remaining models have lost all resemblance to the ground truth. 


\begin{table}[t]
\caption{Relative error of models trained and tested on full trajectories ($t\in[0,5]$).}
\label{metrics_200ts}
\vskip 0.15in
\begin{center}
\begin{small}
\begin{sc}
\begin{tabular}{llcc}
\toprule
\textbf{E1: Interpolation.} & & \\
\midrule
\textbf{Two-step training}   & Relative $L_1$  & Relative $L_2$ \\
\midrule

DeepONet & 0.0892$\pm$0.0088 &    0.0282$\pm$0.0022  \\
DON+RNN  & 0.0573$\pm$0.0032 &    0.0196$\pm$0.0005 \\
DON+GRU  & 0.0429$\pm$0.0028 &    0.0151$\pm$0.0006   \\
DON+LSTM & 0.0396$\pm$0.0005 &      0.0154$\pm$0.0006  \\
\midrule
FNO       & 0.0625$\pm$0.0021 &    0.0076$\pm$0.0008 \\
FNO+RNN   & 0.0425$\pm$0.0013 &     0.0025$\pm$0.0001   \\
FNO+GRU   & 0.0205$\pm$0.0019 &     \textbf{0.0007$\pm$0.0001} \\
FNO+LSTM  & \textbf{0.0193$\pm$0.0010} &     \textbf{0.0007$\pm$0.0001}  \\
\midrule
\textbf{Simultaneous} & &\\
\textbf{training}   & Relative $L_1$ & Relative $L_2$ \\
\midrule
DON-RNN  & 0.0562$\pm$0.0101 &     0.0038$\pm$0.0012   \\
FNO-RNN & 0.0486$\pm$0.0008 & 0.0031$\pm$0.0001  \\

\bottomrule
\end{tabular}
\end{sc}
\end{small}
\end{center}
\vskip -0.1in
\end{table}

%

\begin{table}[ht]
\caption{Relative error of models trained on partial trajectories (50 steps $t\in[0, 1.25]$), tested on the full trajectories in a recursive manner ($200$ steps in $t\in[0,5]$), where the output of the last prediction step is the input in the next prediction).}
\label{metrics_50ts}
\vskip 0.15in
\begin{center}
\begin{small}
\begin{sc}
\begin{tabular}{lccr}
\toprule
\textbf{E2: Extrapolation.} & & \\
\midrule
\textbf{Two-step training}   & Relative $L_1$  & Relative $L_2$ \\
\midrule

DeepONet & 0.1060$\pm$0.0044 &     0.0604$\pm$0.0090  \\
DON+RNN  & 0.1048$\pm$0.0032 &    0.0611$\pm$0.0084   \\
DON+GRU  & 0.1008$\pm$0.0078 &    0.0583$\pm$0.0096   \\
DON+LSTM & 0.1009$\pm$0.0074 &    0.0601$\pm$0.0100    \\
\midrule
FNO       & 0.2399$\pm$0.0243 &    0.2116$\pm$0.0473   \\
FNO+RNN   & 0.1815$\pm$0.0222  &    0.1587$\pm$0.0452    \\
FNO+GRU   & 0.1019$\pm$0.0138 &    0.0705$\pm$0.0222 \\
FNO+LSTM  & 0.0901$\pm$0.0083 &    0.0477$\pm$0.0134  \\
\midrule
\textbf{Simultaneous} & &\\
\textbf{training}   & Relative $L_1$ & Relative $L_2$ \\
\midrule
DON-RNN  &\textbf{ 0.0827$\pm$0.0016} &   \textbf{0.0275$\pm$0.0006}     \\
FNO-RNN & 0.1970$\pm$0.0104  & 0.1371$\pm$0.0297  \\

\bottomrule
\end{tabular}
\end{sc}
\end{small}
\end{center}
\vskip -0.1in
\end{table}


\section{Conclusions}

While neural methods have been shown to be promising surrogate models to accelerate the simulation of dynamical systems, integrating these systems over long-time horizons still remains an open challenge. 
Some recent works propose to address this problem by employing physics-informed deep operator networks, transfer learning, or hybrid approaches, however, these methods suffer from their own limitations.

In this work, we have proposed to tackle extrapolation and long-time horizon prediction through multiple variants combining neural operators with recurrent neural networks. The resulting models show improvement in terms of accuracy and stability of prediction compared to their vanilla counterparts.



Our observations from this study can be summarized as follows:
\begin{enumerate}
    \item The explored extension of the neural operators reduces the overall error in both interpolation and extrapolation experiments when compared to vanilla models, as well as the error increase over the whole trajectory for all models except for the extrapolation on simultaneously trained FNO-RNN (Tables~\ref{metrics_200ts} and \ref{metrics_50ts}, Figures~\ref{fig:interpolation_performance} and \ref{fig:extrapolation_performance}).
    \item The error growth rate is also reduced for most of the models. In the interpolation tests, the accumulated error practically flattens out for all extensions of neural operators, with the exception of the models trained with a DeepONet in the two-step process, where the error is lower, but still shows significant increase (\autoref{fig:interpolation_performance}). In extrapolation, the reduction in the error growth rate is significant for the FNO trained in a two-step process, as well as for both operator-RNN architectures trained simultaneously  (Figure~\ref{fig:extrapolation_performance}a).
    \item The combined architecture is shown to improve the ability of the operator to maintain the shape of the solution, which prevents some of the error propagation in long-time integration (Figures~\ref{fig:200ts} and \ref{fig:50ts}). The stability is more pronounced when the components of the integrated framework are trained simultaneously.
\end{enumerate}


Despite the promising results presented here, we must acknowledge that using the RNN-integrated neural operator architecture to solve long-term prediction problems is still in its infancy. Numerous open issues should be taken into account as potential future study topics. It is crucial from a theoretical standpoint to gain a better understanding of how approximation errors impact the stability and precision of the suggested methods. This is a key component in critical applications where accuracy and convergence guarantees are required, where conventional numerical solvers are still the default option at the moment. 

\bibliographystyle{IEEEtran}
\bibliography{main}
\appendix{}
\subsection{Evaluation metrics}
\label{app:eval_metrics_explained}

The performance of the presented models is expressed in relative error. Relative error values range from $0$ to infinity, where 0 represents a perfect prediction, and values above $1$ indicate predictions worse than the mean of the actual values as a prediction.

The relative $L_1$ error is given as:
\begin{equation}
    L_1 = 
    \frac{
        \frac{1}{n}\sum_{i=1}^{n}(y_i - \hat{y_i})^2
        }
        {
        \frac{1}{n} \sum_{i=1}^{n}(y_i - \bar{y})^2
        }
\end{equation}
where $y_i$ denotes the actual value for the $i^{th}$ sample, $\hat{y_i}$ is the predicted value for the $i^{th}$ sample, and $\bar{y}$ is the average of all actual values.

The relative $L_2$ error is given as:
\begin{equation}
    L_2 = 
    \frac{
        \frac{1}{n}\sum_{i=1}^{n}(y_i - \hat{y_i})^2
        }
        {
        \frac{1}{n} \sum_{i=1}^{n}(y_i - \bar{y})^2
        }
\end{equation}
where $y_i$ denotes the actual value for the $i^{th}$ sample, $\hat{y_i}$ is the predicted value for the $i^{th}$ sample, and $\bar{y}$ is the average of all actual values.

\section{Architecture details}

\label{app:architectures} 

\subsection{Network architectures}

\begin{table}[th]
\caption{DeepONet/DON-RNN size. The spatial size $x=50$ and the temporal size $t=200$. Batch size is represented as -1.}
\label{don_branch_architecture}
\vskip 0.15in
\begin{center}
\begin{small}
\begin{sc}
\begin{tabular}{llcr}
\toprule
& Input & Layer sizes & Output    \\
\midrule
Branch & [-1, $x$]  & 150, 250, 450, 380, 320 &   [-1, 300]  \\
Trunk & [-1, 2] & 200,220,240,250,260,280 & [-1, 300]  \\
RNN & [-1, $xt$] & 200, $x$ & [-1, $t$, $x$]\\
\bottomrule
\end{tabular}
\end{sc}
\end{small}
\end{center}
\vskip -0.1in
\end{table}

\begin{table}[th]
\caption{Fourier neural operator architecture for the input of 200 timesteps. Total nr of parameters is $25,345$.}
\vskip 0.15in
\begin{center}
\begin{small}
\begin{sc}
\begin{tabular}{llccr}
\toprule
 & Layer & Layer sizes & Activation & Param   \\
\midrule
0   & Input &   [-1, 200, 50, 1] & - & $0$ \\
1   & Dense &   [-1, 200, 50, 64] & linear & $256$ \\
2   & Fourier &  [-1, 64, 209, 59] & - & $4,160$ \\
3   & Fourier &  [-1, 64, 209, 59] & - & $4,160$ \\
4   & Fourier &  [-1, 64, 209, 59] & - & $4,160$ \\
5   & Fourier &  [-1, 64, 209, 59] & - & $4,160$ \\
6   & Dense &  [-1, 200, 50, 128] & linear & $8,320$ \\
7   & Dense &  [-1, 200, 50, 1] & linear & $129$ \\
\bottomrule
\end{tabular}
\end{sc}
\end{small}
\end{center}
\vskip -0.1in
\end{table}


The DeepONet is composed of a branch and trunk network with same-size outputs, which are combined by element-wise multiplication, resulting in the output shaped [-1, $xt$], where $x$ is the spatial dimension and $t$ is the temporal dimension. 

The branch and trunk networks employ swish activation and the simple RNN, GRU and LSTM networks use tanh activation. The training is performed in minibatches of size $256$ with Adam optimizer and learning rate of $1e-4$. The inputs to the branch network and the outputs of the element-wise multiplication later are normalized with a standard scaling, and the inputs to the trunk network use min-max scaling. The inputs and outputs in the two-step RNN training are normalized with standard scaler. The vanilla DeepONet and the DON-RNN are trained up to $20,000$ epochs and the RNN in the two-step training is trained for $5,000$ epochs, as no improvement is seen for longer training. 

The total number of parameters of the branch network is $547,000$ and of the trunk network $380,750$. The RNN extension adds $60,250$ parameters. When using gated units instead of the simple RNN, the number of parameters is considerably higher: $161,250$ for GRU, and $210,850$ for LSTM.


A single Fourier layer is constructed with a 2D spectral convolution and a 2D-convolution skip connection.

The FNO is trained in minibatches of 50 samples using the Adam optimizer, the learning rate of $1e-3$ and a scheduler with a step size of 100. The inputs and outputs of the FNO are not scaled. The vanilla FNOs are trained up to $5000$ epochs, while FNO-RNN up to $2000$ epochs. The models used in inference are the ones at a $1000$ epochs, as no further improvement is seen for longer training.

\end{document}